\documentclass[sigconf]{acmart}

\usepackage{color}
\usepackage{soul}
\usepackage{url}
\usepackage[utf8]{inputenc}
\usepackage{graphicx}
\usepackage{amsmath}
\usepackage{booktabs}
\usepackage{amssymb}
\usepackage{enumerate}
\usepackage{multirow}
\usepackage{caption}
\usepackage{algorithm}
\usepackage{algorithmic}
\copyrightyear{2020}
\acmYear{2020}
\setcopyright{acmcopyright}
\acmConference[MM '20] {28th ACM International Conference on Multimedia}{October 12--16, 2020}{Seattle, WA, USA.}
\acmBooktitle{28th ACM International Conference on Multimedia (MM '20), October 12--16, 2020, Seattle, WA, USA.}
\acmPrice{15.00}
\acmDOI{10.1145/3394171.3413968}
\acmISBN{978-1-4503-7988-5/20/10}

\begin{document}
\fancyhead{}

\title{Discernible Image Compression}

\author{Zhaohui Yang$^{1, 2}$, Yunhe Wang$^{2}$, Chang Xu$^3\dagger$ , Peng Du$^4$, Chao Xu$^{1}$, Chunjing Xu$^{2}$, Qi Tian$^5$}
\authornote{$\dagger$ corresponding author.}
\affiliation{%
  \institution{$^1$ Key Lab of Machine Perception (MOE), Dept. of Machine Intelligence, Peking University. \\
  $^2$ Noah’s Ark Lab, Huawei Technologies. \\
  $^3$ School of Computer Science, Faculty of Engineering, University of Sydney. \\
  $^4$ Huawei Technologies. $^5$ Cloud \& AI, Huawei Technologies. \\
  }
}
\email{zhaohuiyang@pku.edu.cn, {yunhe.wang, xuchunjing, tian.qi1}@huawei.com, c.xu@sydney.edu.au}
\email{dp@zju.edu.cn, xuchao@cis.pku.edu.cn }

\newcommand{\eg}{\emph{e.g.}}
\newcommand{\etal}{\emph{et.al.}}
\newcommand{\ie}{\emph{i.e.}}
\newcommand{\etc}{\emph{etc}}
\newcommand{\wrt}{\emph{w.r.t. }}
\newcommand{\iid}{\emph{i.i.d. }}

\begin{abstract}
Image compression, as one of the fundamental low-level image processing tasks, is very essential for computer vision. Tremendous computing and storage resources can be preserved with a trivial amount of visual information. Conventional image compression methods tend to obtain compressed images by minimizing their appearance discrepancy with the corresponding original images, but pay little attention to their efficacy in downstream perception tasks, \eg, image recognition and object detection. Thus, some of compressed images could be recognized with bias. In contrast, this paper aims to produce compressed images by pursuing both appearance and perceptual consistency. Based on the encoder-decoder framework, we propose using a pre-trained CNN to extract features of the original and compressed images, and making them similar. Thus the compressed images are discernible to subsequent tasks, and we name our method as Discernible Image Compression (DIC). In addition, the maximum mean discrepancy (MMD) is employed to minimize the difference between feature distributions. The resulting compression network can generate images with high image quality and preserve the consistent perception in the feature domain, so that these images can be well recognized by pre-trained machine learning models. Experiments on benchmarks demonstrate that images compressed by using the proposed method can also be well recognized by subsequent visual recognition and detection models. For instance, the mAP value of compressed images by DIC is about 0.6\% higher than that of using compressed images by conventional methods.
\end{abstract}



\maketitle


\section{Introduction}
Recently, more and more computer vision (CV) tasks such as image recognition~\cite{ResNet,cars,hournas,dafl,legonet}, visual segmentation~\cite{FCN}, object detection~\cite{RCNN,fasterRCNN,hitdetector}, and face verification~\cite{DeepID2}, are well addressed by deep neural networks, which benefits from the large amount of accessible training data and computational power of GPUs. Besides these high-level CV tasks, a lot of low-level CV tasks have been enhanced by neural networks, \eg, image denoising and inpainting~\cite{denoise1,BeyondGaussian,denoise2}, single image super-resolution~\cite{SRCNN,song2020efficient}, and image and video compression~\cite{lossy1,RNN2,lossy5}.

This paper studies the image compression problem, a fundamental approach for saving storage and transmission consumptions, which represents images with low-bit data and reconstructs them with high image quality. Traditional methods are mainly based on the time-frequency domain transform (\eg, JPEG~\cite{JPEG} and JPEG 2000~\cite{JPEG2000}), which makes compressed images distorted with blocking artifacts or noises. Since convolutional networks have shown extraordinary performance on image denoising and inpainting~\cite{denoise1,denoise2}, Dong \emph{et al.}~\cite{BeyondGaussian} and Zhang \emph{et al.}~\cite{deblock1} proposed using CNN to remove blocks on JPEG compressed images in order to enhance the compression performance. Moreover, Toderici \emph{et al.}~\cite{RNN1} utilized an encoder-decoder network to implement the compressing task with a fixed input size. Toderici \emph{et al.}~\cite{RNN2} further extended the encoder-decoder network to a general model that supports images with arbitrary sizes. Sun~\emph{et al.}~\cite{aaai2} utilized the recursive dilated network to establish the image compression system. Li \emph{et al.}~\cite{lossy3} proposed to compress images by exploiting the importance map to achieve higher compression rates.

\begin{figure*}[t]
\begin{center}
\begin{tabular}{cccc}
\emph{\color{blue} Pilecan}, $20.6\%$ & \emph{\color{red} Dowitcher}, $21.0\%$ & \emph{\color{red} Mergus Serrator}, $16.6\%$ & \emph{\color{red} Dunlin}, $20.7\%$ \\
\includegraphics[width=0.23\linewidth]{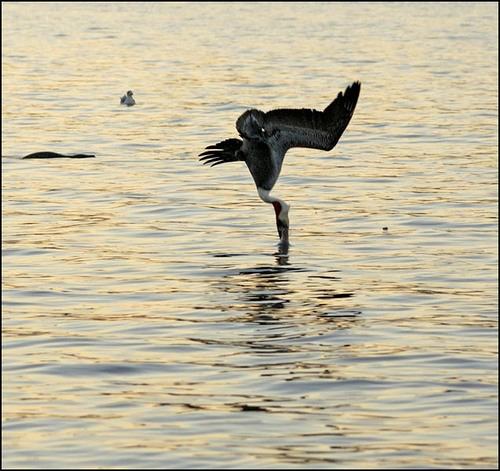}&\includegraphics[width=0.23\linewidth]{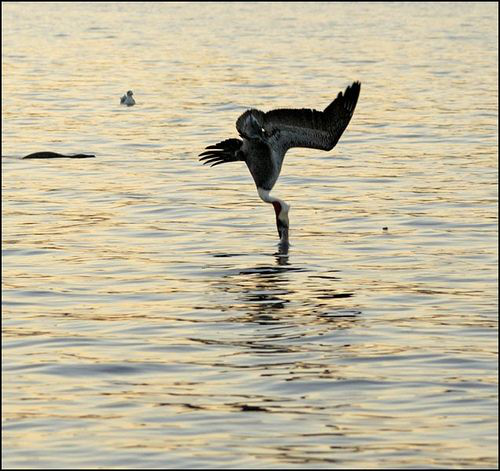}&\includegraphics[width=0.23\linewidth]{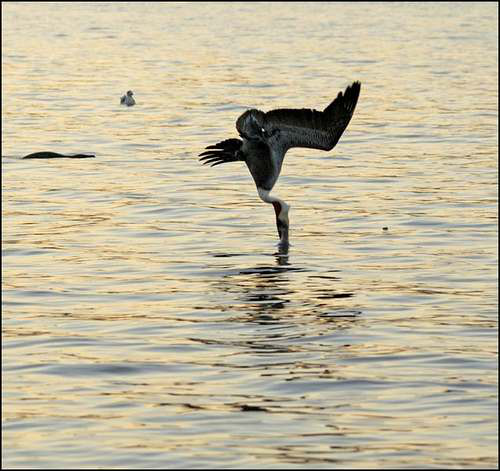}&\includegraphics[width=0.23\linewidth]{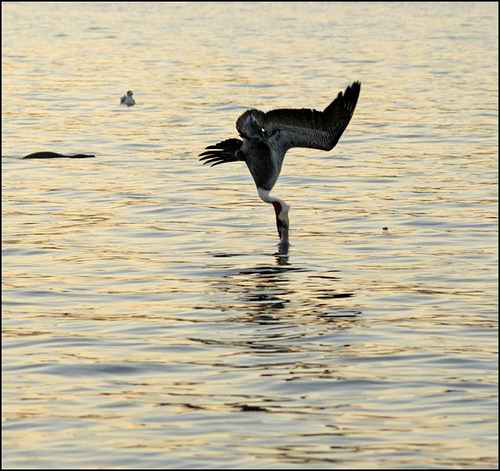} 
\end{tabular}
\caption{Recognition results of compressed images using a pre-trained ResNet on an image labeled as \emph{\color{blue} Pilecan}. From left to right are compressed images, and the $bpp = 2.5, 1.8, 1.2$, and $1.0$, respectively. Recognition results (label and score) are shown at the top of images. }
\label{Fig:Intro}
\end{center}
\end{figure*}

Although these methods obtained promising performance to reduce the storage of digital images, there is an important issue that should be considered. In practice, a number of digital images will be taken and stored in electronic devices (\eg, mobile phones, security cameras), and a large proportion of them will be recognized or post-processed using pre-trained machine models for person identification, object recognition and detection, \etc. Therefore, we do not expect that compressed images cannot be accurately recognized by these pre-trained models after compressing. However, the big success of the deep neural network is mainly contributed to the massive amount of available data. Thus, almost all widely used neural networks are sensitive to small changes in the given images. Some different texture and color changes outside the original training dataset will directly interfere with the output results. In addition, since the network for compressing images is trained for minimizing the pixel-wise error between any two images, which cannot be integrated with subsequent recognition tasks well.

To have an explicit illustration, we conduct a toy experiment as shown in Figure~\ref{Fig:Intro}. In practice, we directly compress an image by using the JPEG algorithm and recognize the compressed image using a pre-trained ResNet-50~\cite{ResNet}. The recognition results, \emph{i.e.}, the score of the predicted score will be changed, while the compression rate is continuously increasing. Although there is only a very small appearance difference between the original image and the compressed image, some underlying structure and textual changes will affect the calculation of the subsequent neural network. The network recognizes some compressed \emph{\color{blue}Pilecan} images as \emph{\color{red}Dowitcher, Mergus Serrator}, and {\color{red} Dunlin}, respectively, though there is no hurdle for use to recognize them by eyes. On the other side, although we could retrain existing models (\eg, classification or detection) for fitting these compressed images, the time consumption is not tolerable. Another attempt could be adding more compressed images as augmentations, but there is a number of image compression algorithms and different sizes of images. Therefore, an image compression method for generating compressed images with perception consistency in the feature domain is urgently required.

To address the aforementioned problems, this paper develops a novel image compression framework that simultaneously executes image compression and image recognition tasks. In specific, an encoder-decoder network is used for generating compressed data and reconstructed images, and a pre-trained CNN is adopted to perceive the difference of images after and before compression. By jointly optimizing these two objectives, the proposed method can produce compressed images with low storage which can also be accurately discerned as usual by pre-trained CNN models. To the best of our knowledge, this is the first time to simultaneously investigate the physical compression and visual perception of images using deep learning methods. Experiments conducted on benchmark datasets demonstrate the superiority of the proposed algorithm over the state-of-the-art methods for compressing digital images.

\section{Related Works}

\begin{figure*}[t]
	\begin{center}
		\includegraphics[width=\linewidth]{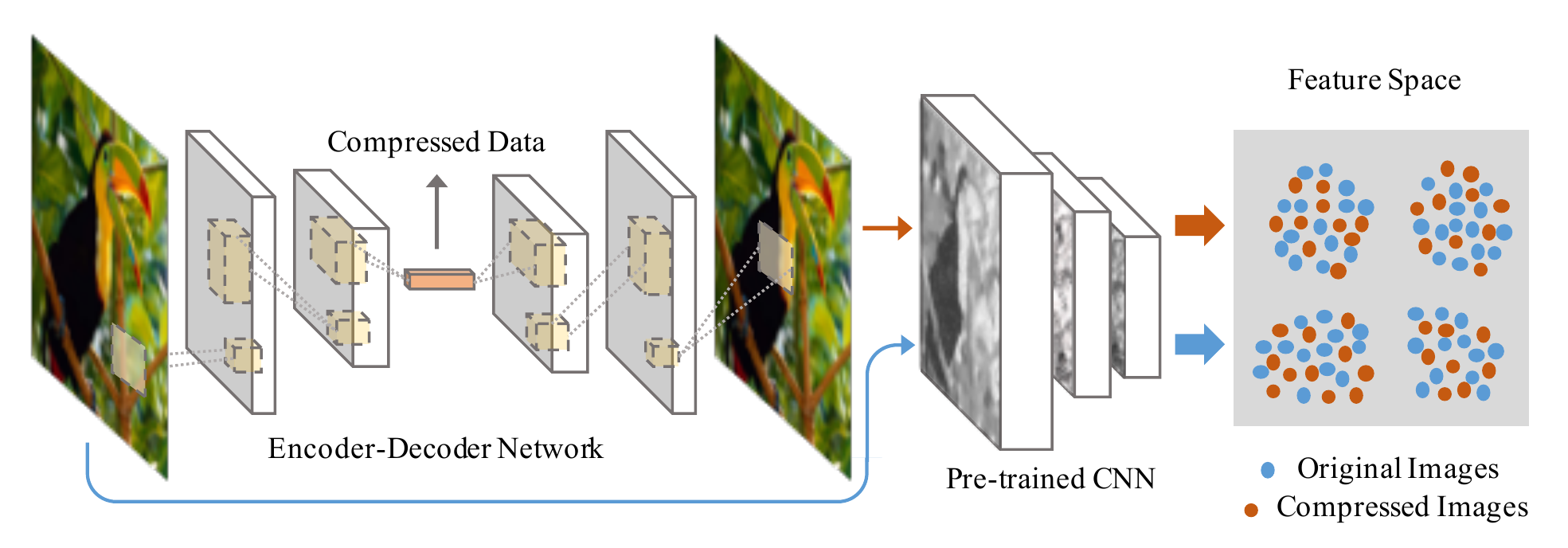}
	\end{center}
	\caption{The diagram of the proposed method for compressing digital images, which consists of an encoder-decoder network and a pre-trained convolutional neural network. Wherein, the encoder-decoder network represents images with limited storage and the subsequent CNN is used for extracting features of all images simultaneously. Features of compressed images will then align to those of original images by minimizing their distances and distributions.}
	\label{Fig:diagram}
\end{figure*}

Deep learning methods have shown extraordinary performance on a variety of image processing tasks such as image denoising~\cite{BeyondGaussian} and super-resolution~\cite{SRCNN}, which can provide clear and smooth outputs. Therefore, it is meaningful to devise end-to-end image compression methods based on deep models. 

For the lossless image compression,~\cite{lossless1} and~\cite{lossless2} proposed image compression methods that can preserve all image information, but their compression rates are not competitive. On another side, compared with traditional methods which have several independent components such as quantization, pruning, and encoding~\cite{JPEG}, the deep learning method using an end-to-end strategy is more effective. However, the main difficulty is that the rounding or binary function is non-differentiable. Theis~\etal~\cite{lossy1} proposed compressive auto-encoders, which develops a smooth approximation of the discrete of the rounding function. Jiang~\etal~\cite{lossy4} generated low-resolution images and then applied conventional image compression methods such as JPEG to obtain low-bit data. Cheng~\etal~\cite{deepresiduallearningforimagecompression} investigated residual learning for image compression. Akyazi~\etal~\cite{waveletdecomposition} used the wavelet decomposition for learning. Murashko \etal~\cite{murashko2016predicting} proposed an ml-based method to model the outcome of an arbitrary image. The image compression method were integrated with other computer vision tasks and performed well. For example, Jia~\etal~\cite{comdefend} proposed an end-to-end image compression model composed of ComCNN and ResCNN to defend adversarial examples.

A number of works proposed novel compression schemes to improve the quality of compressed images. Savioli~\cite{hybridapproach} combines RL and GAN to maximum PSNR for image compression. Rippel and Bourdev~\cite{lossy5} developed an efficient image compression system with auto-encoder and adversarial training. Choi~\etal~\cite{variableratedeepimagecompression} developed a variable-rate image compression model with a conditional autoencoder. Cai~\etal~\cite{cai2019efficient} integrated the variational auto-encoder architecture with sub-pixel image compression. Based on convolutional and de-convolutional LSTM recurrent networks (RNN), Lee~\etal~\cite{contextadaptiveentropymodel} explored bit-consuming and bit-free contexts and build models based on the contexts. Cui \etal~\cite{cui2018an} proposed novel deep learning-based compressed sensing coding framework. Li~\etal~\cite{lossy3} proposed an attention-based image compression approach that learns the importance map and the compressed data simultaneously. 

The encoder/decoder, quantization function, entropy coding and loss function are the basic components for training an image compressor. Optimizing each component could improve the global compressor performance. Toderici~\etal~\cite{RNN1} proposed a framework for variable-rate image compression. Toderici~\etal~\cite{RNN2} further expanded the RNN based method to a general approach that is competitive across compression rates on arbitrary images with different sizes. Liu~\etal~\cite{liu2018deepn} found the JPEG does not fit for the neural network system and designed DeepN-JPEG which reduce the quantization error. Ball{\'e}~\etal~\cite{lossy2} utilized a generalized divisive normalization to replace rounding quantization by additive uniform noise for continuous relaxation. Mentzer~\etal~\cite{practicalfullresolution} explored a probabilistic model for end-to-end adaptive entropy coding. Aytekin~\etal~\cite{cycleloss} proposed cycle loss to train image compression models. Cheng~\etal~\cite{spatialtemporalenergycompaction} proposed the spatial energy compaction-based penalty to enhance performance. 

Although the aforementioned image compression methods make tremendous efforts to learn deep learning models for compressing images, most of them only minimize the appearance difference between the given image and the compressed image, but ignore the perception difference of these images in the feature domain. Hence, they often lead to the unexpected failure of visual perception, as shown in Figure1 in the main body. We aim to explore an image compression approach that can preserve appearance and perception consistency of compressed image simultaneously. 
 
\section{Perception-consistent Image Compression}\label{Sec:Model}
The encoder-decoder network receives and outputs images and represents them with activations of a number of hidden neurons, which is naturally suitable for implementing the image compression task~\cite{lossy1,RNN1,RNN2}.

\subsection{Encoder-decoder for Image Compression}
Generally, the loss function of an encoder-decoder based image compression network can be written as
\begin{equation}
\min_{\theta_1,\theta_2}\frac{1}{n}\sum_{i=1}^n||\mathcal{D}\left(\theta_2,\mathcal{E}(\theta_1,x^i)\right)-x^i||^2,
\label{Fcn:enco}
\end{equation}
where $\mathcal{E}(\cdot)$ is the encoder network with parameter $\theta_1$ for compressing the given image $x^i$, $n$ is the number of images, and $\mathcal{D}(\cdot)$ is the decoder network with parameter $\theta_2$ for recovering the compressed data to the original image. To simplify the expression, the compressed data $c$ and decoded image $y$ are denoted as
\begin{equation}
c^i = \mathcal{E}(\theta_1,x^i),\quad y^i=\mathcal{D}\left(\theta_2,c^i\right),
\end{equation}
respectively. Since the encoder network $\mathcal{E}$ consists of a series of transforms such as convolution, pooling, binary (or quantization) and pruning, the decoded image $y^i$ resulted from conventional JPEG algorithm~\cite{JPEG} often has some distortions such as blocks and artifacts. Therefore, a feasible way for enhancing the image quality of $y^i$ is to use another operation or model for refining the decoded image $y^i$:
\begin{equation}
\min_{\hat{y}} \frac{1}{n}\sum_{i=1}^n\left(||\hat{y}^i-y^i||^2+\lambda \mathcal{R}(\hat{y}^i)\right),
\label{Fcn:deblock}
\end{equation}
where $\hat{y}^i$ is the recovered image and $\mathcal{R}(\cdot)$ can be some conventional regularizations for natural images such as total variation (TV) norm, $\ell_1$ norm, \etc. These techniques have been widely used in conventional image processing methods~\cite{WNNM,NCSR}. In contrast, neural networks of strong capacity can generate clearer images than those of traditional methods. Additional layers can be easily added after $\mathcal{D}(\cdot)$ for refining $y$, so that the above function can be absorbed by Fcn.~\ref{Fcn:enco} to form an end-to-end model learning framework.

An ideal image compression algorithm should not only concentrate on the low storage of generated images, but also have to retain the downstream performance of tasks such as image recognition and object detection, \etc. Therefore, a neural network with parameter $\theta_3$ is introduced for supervising the generated image:
\begin{equation}
\begin{aligned}
\min_{\theta_1,\theta_2,\theta_3}&\frac{1}{n}\sum_{i=1}^n\left(||y^i-x^i||^2 + \lambda\mathcal{L}(y^i,\theta_3)\right),\\
\quad &s.t.\;\; y^i = \mathcal{D}\left(\theta_2,\mathcal{E}(\theta_1,x^i)\right),
\end{aligned}
\label{Fcn:test}
\end{equation}
where $\mathcal{L}(\cdot)$ can be various based on different applications, \eg, cross-entropy loss, regression loss. 

There are a number of off-the-shelf visual models (\eg, ResNet~\cite{ResNet}, VGGNet~\cite{VGGNet}, RCNN~\cite{RCNN}) well trained on large-scale datasets (\eg, ILSVRC~\cite{ImageNet} and COCO~\cite{COCO}), and completely retraining them is a serious waste of resources and is time-consuming. In addition, most existing deep learning models only support input images with fixed sizes (\eg, $224\times244$), while a complete image compression system should be used for processing images with different sizes. 

Hence, we propose to use a pre-trained neural network $\mathcal{F}(\cdot)$ as a ``perceptron'' to the process of original and compressed images simultaneously. Therefore, we formulate a novel image compression method giving consideration to both appearance and perception consistency:
\begin{equation}
\begin{aligned}
\min_{\theta_1,\theta_2}\frac{1}{n}\sum_{i=1}^n&\left(||y^i-x^i||^2 + \lambda ||\mathcal{F}(y^i)-\mathcal{F}(x^i)||^2\right),\\
\quad &s.t.\;\; y^i = \mathcal{D}\left(\theta_2,\mathcal{E}(\theta_1,x^i)\right),
\end{aligned}
\label{Fcn:Feat}
\end{equation}
where $\lambda$ is the trade-off parameter, and parameters in $\mathcal{F}(\cdot)$ are fixed. The diagram of the proposed network is shown in Figure~\ref{Fig:diagram}. Since the size of the decoded image can be varied according to different sizes of the input image, $\mathcal{F}(\cdot)$ cannot be the same as the original pre-trained network. In practice, $\mathcal{F}(\cdot)$ is a pre-trained CNN after discarding the last several layers, which can generate features with different dimensionalities given different images. Although Fcn.~\ref{Fcn:Feat} does not explicitly optimize the recognition performance, considerable convolution filters in $\mathcal{F}(\cdot)$ trained over a number of images can be beneficial for perceiving. Therefore, minimizing Fcn.~\ref{Fcn:Feat} will encourage the perception consistency between compressed image $y$ and its corresponding original image. 

\subsection{Feature Distribution Optimization}
A novel image compression model was proposed in Fcn.~\ref{Fcn:Feat}, which introduces a new module $\mathcal{F}(\cdot)$ for extracting visual features of original and compressed images. Since there are considerable neurons in a well-designed neural network, $\mathcal{F}(\cdot)$ will convert input images into high-dimensional (\eg, $512$) features, and it is very hard to minimize differences between features of these images directly. Therefore, we propose to use another measurement to supervise the compression task, \ie, \emph{maximum mean discrepancy} (MMD~\cite{MMD,MKMMD}), which is used for describing differences of two distributions by mapping sample data in kernel spaces.

Suppose we are given an image dataset with $n$ images, and two sets of image features $\mathcal{X} = \{\mathcal{F}(x^i)\}_{i=1}^{n}$ sampled from distribution $p$ and $\mathcal{Y} = \{\mathcal{F}(y^i)\}_{i=1}^{n}$ sampled from distribution $q$, where $x^i$ and $y^i$ are the $i$-th original image and the $i$-th compressed image, respectively. The squared formulation of MMD distance between $p$ and $q$ is defined as:
\begin{equation}
\small
\mathcal{L}_{MMD}(\mathcal{X},\mathcal{Y}) \triangleq \frac{1}{n}||\sum_{i=1}^n\psi(\mathcal{F}(x^i))-\sum_{i=1}^n\psi(\mathcal{F}(y^i)) ||^2,
\end{equation}
where $\psi(\cdot)$ is an explicit mapping function. It is clear that feature distributions of original and compressed images are exactly the same iff $\mathcal{L}_{MMD}=0$, \ie, $p=q$~\cite{MMD}. The above function can be further expanded with the kernel trick:
\begin{equation*}
\footnotesize
\begin{aligned}
\mathcal{L}&_{MMD}(\mathcal{X},\mathcal{Y}) = \frac{1}{n^2}\left[\sum_{i=1}^n\sum_{i'=1}^nk(\mathcal{F}(x^i),\mathcal{F}(x^{i'}))+ \right.\\
&\left. \sum_{i=1}^n\sum_{i'=1}^nk(\mathcal{F}(y^i),\mathcal{F}(y^{i'}))-\sum_{i=1}^n\sum_{j=1}^nk(\mathcal{F}(x^i),\mathcal{F}(y^{j}))\right],
\end{aligned}
\end{equation*}
where $k(\cdot,\cdot)$ is a kernel function for projecting given data into a higher or infinite dimensional space, which can be set as a linear kernel, Gaussian kernel, \etc. Since each kernel has its own functionality for measuring distributions of data, it is very hard to determine which one is the best in practice without time consuming cross-validation. Therefore, we borrow the strategy in~\cite{MKMMD} to use a set of kernels for projecting features:
\begin{equation*}
\footnotesize
\left\{ k=\sum_{u=1}^m\beta_uk_u :\sum_{u=1}^m\beta_u=1, \beta_u\geq0, \forall u \right\},
\end{equation*}
where $m$ is the number of kernels, $\beta_u$ is the coefficient of the $u$-th kernel which can be optimized iteratively. Therefore, we reformulate Fcn.~\ref{Fcn:Feat} as
\begin{equation}
\begin{aligned}
\mathcal{L}_{Comp}(&\theta_1,\theta_2)=\frac{1}{n} \sum_{i=1}^n||y^i-x^i||^2+\\
 \frac{\lambda}{n}&\sum_{i=1}^n||\mathcal{F}(y^i)-\mathcal{F}(x^i)||^2+\gamma \mathcal{L}_{MMD}(\mathcal{X},\mathcal{Y}),\\
&\quad\quad s.t.\;\; y^i = \mathcal{D}\left(\theta_2,\mathcal{E}(\theta_1,x^i)\right), \\
&\; \mathcal{X} = \{\mathcal{F}(x^i)\}_{i=1}^{n}, \mathcal{Y} = \{\mathcal{F}(y^i)\}_{i=1}^{n},
\end{aligned}
\label{Fcn:obj}
\end{equation}
where $\gamma$ is the weight parameter for the MMD loss. By simultaneously optimizing the compression loss and the perception loss, we can obtain a model which generates compressed images of the consistent perception with original images for a series of downstream tasks such as image recognition and segmentation, \etc. Alg.~\ref{Alg:main} summarizes the mini-batch strategy of the proposed method for the learning image compression network. In addition, the pre-trained network $\mathcal{F}(\cdot)$ will be dropped after the training process.

\textbf{Discussion.} The proposed method includes the image compression task and the visual perception task. Wherein, a pre-trained neural network is utilized for extracting features of original and compressed images, which is similar to two categories of works, \ie, transfer learning~\cite{MKMMD} and teacher-student learning paradigm~\cite{Distill,FitNet}, which also utilize a pre-trained model for inheriting useful information for helping the training process. The main difference is that we do not train any new parameters for the visual recognition task, and parameters in the pre-trained network are fixed, which is used as a powerful regularization for supervising the learning of the encoder-decoder network therefore improves the compression performance.

\begin{algorithm}[t]
\caption{Discernible Images Compression Method.}
\label{Alg:main}
\begin{algorithmic}[1]
\REQUIRE An image datasets $\{x^1,...,x^n\}$ with $n$ images, a pre-trained CNN $\mathcal{F}(\cdot)$, $\lambda$, $\gamma$, and batch size $b$.
\STATE Initialize the compression network with an encoder $\mathcal{E}(\cdot)$ and a decoder $\mathcal{D}(\cdot)$ and their parameters $\theta_1$ and $\theta_2$, respectively.
\REPEAT
\STATE Randomly select a batch of $b$ image $\{x^1,...,x^b\}$;
\FOR{$i=1$ to $b$}
\STATE Compress the given image $c^i \leftarrow \mathcal{E}(x^i)$;
\STATE Decode the compressed data $y^i \leftarrow \mathcal{D}(c^i)$; 
\STATE Generate image features $\mathcal{F}(x^i)$ and $\mathcal{F}(y^i)$;
\ENDFOR
\STATE Calculate $\mathcal{L}_{Comp}(\theta_1,\theta_2)$ according to Fcn.~\ref{Fcn:obj};
\STATE Update $\theta_1$ and $\theta_2$ according to $\mathcal{L}_{Comp}(\theta_1,\theta_2)$;
\UNTIL convergence
\ENSURE The optimal compression model with $\mathcal{E}$ and $\mathcal{D}$.
\end{algorithmic}
\end{algorithm}

\section{Experiments}\label{Sec:Exp}

We conduct the experiments on several datasets to demonstrate the effectiveness of the proposed perceptual consistency.

\subsection{Datasets}

Three widely used datasets are selected to conduct the visual recognition and detection tasks, \ie, ImageNet~\cite{ImageNet}, COCO~\cite{COCO}, and VOC~\cite{VOC}. Most of the datasets used for recognition and detection tasks are provided of the JPEG format, and the average image size is around 100KB. The provided images are quite large, thus we view the images as the original, and they are used for calculating MS-SSIM and PSNR.

\subsubsection{Image Recognition}

\paragraph{ImageNet} The ImageNet 2012 dataset~\cite{ImageNet} is a large scale recognition dataset. The ImageNet 2012 train set contains 1.28 million images, and the validation set contains 50,000 images. We use the models trained on the ImageNet train set to extract features for calculating the proposed perceptual losses. Besides, after training the image compression models, the models are used to compress the ImageNet 2012 validation set, and the compressed images are to be evaluated for the subsequent recognition task. 

\subsubsection{Object Detection}

\paragraph{COCO} The MS-COCO dataset~\cite{COCO} is a challenging dataset for detection and segmentation tasks. The dataset contains 1.5 million annotated instances which belong to 80 categories. We follow the baseline method~\cite{RNN2} and use the COCO trainval set to train the image compression models.

\paragraph{VOC} The VOC 2007 dataset~\cite{VOC} includes 9,963 images split into trainval/test sets which separately include 5,011 and 4,952 images. We use the trained image compression models to compress the test set and use the pretrained detection models~\cite{SSD} to measure the mAP.

\subsection{Implementation Details}

\paragraph{Baseline Model} There are a number of CNN based image compression models~\cite{lossless1,lossy1,aaai1,aaai2,RNN2,baig2017learning,CPM}, each of which has its own pros and cons. We selected~\cite{RNN2} as the baseline model, which utilizes a recurrent neural network (RNN) for compressing images for the following two reasons: 1) the RNN based encoder-decoder network can provide compressed images with different compression rates in each iteration; 2) this model allows the size of the input image to be arbitrary, which is more flexible than comparison methods with fixed input size.

\begin{table}[t]
	\caption{Recognition results on the ILSVRC 2012 dataset with different hyper-parameters $\lambda$.}
	\label{Tab:comp1}
	\begin{center}
		\begin{tabular}{c|c|c|c|c}
			\hline
			\multirow{2}{1cm}{\centering Method} & \multicolumn{2}{c|}{ResNet-18} & \multirow{2}{1.5cm}{\centering MS-SSIM} & \multirow{2}{1cm}{\centering PSNR}\\	
			\cline{2-3}
			&top-1 & top-5 & \\
			\hline\hline
			Original & $69.8\%$ &$89.1\%$& $1.000$ & - \\
			\hline
			FRIC & $63.5\%$ & $85.0\%$ & 0.921 & 24.442 \\
			\hline
			$\lambda = 10^{-4}$& $63.3\%$ & $84.8\%$ & 0.917 & 24.433 \\
			\hline
			$\lambda = 10^{-5}$& $63.8\%$ & $85.2\%$ & 0.922 & 24.453 \\
			\hline
			$\lambda = 10^{-6}$& $64.0\%$ & $85.4\%$ & 0.925 & 24.465 \\
			\hline
			$\lambda = 10^{-7}$& $63.9\%$ & $85.3\%$ & 0.924 & 24.460 \\
			\hline
		\end{tabular}
	\end{center}
\end{table}

\paragraph{Training Setup} To have a fair comparison, we follow the setting in~\cite{RNN2} to conduct the image compression experiment. Each image in the training dataset was decomposed into $32\times32$ non-overlapping patches and we train 200 epochs in total. The architecture of RNN used in the following experiments is the same as that described in~\cite{RNN2}, and networks were trained using the PyTorch toolbox. As for the subsequent CNN $\mathcal{F}(\cdot)$ (see Fcn.~\ref{Fcn:obj}) for extracting visual features of original and compressed images, we used the ResNet-18 network~\cite{ResNet} which shows excellent performance on visual recognition tasks (\eg, an $89.1\%$ top-5 accuracy and an $69.8\%$ top-1 accuracy on the ILSVRC 2012 dataset with 1000 different labels). All parameters in this network were pre-trained on the ImageNet dataset and will be fixed in the following experiments. Note that, $\mathcal{F}(\cdot)$ used here consists of the first 14 convolutional layers in ResNet-18, since the size of the input image is much smaller than that in the original ResNet-18. In specific, for a given input image size of $32\times32$,  $\mathcal{F}(\cdot)$ outputs a $512$-dimensional feature.

\paragraph{Evaluation Metrics} Peak signal to noise ratio (PSNR) and structural similarity (SSIM) are two widely used criteria for evaluating image quality by comparing original images with compressed images. However, the PSNR only measures the mean square error between the compressed image and its original one, and SSIM ignores image differences in different scales. According to~\cite{RNN2}, besides PSNR and SSIM, we also employ the multi-scale structural similarity (MS-SSIM)~\cite{MSSSIM} for evaluating the performance of the proposed image compression algorithm. The MS-SSIM is applied on each RGB channel and we averaged them as the evaluation result. In addition, MS-SSIM values are between 0 and 1. For a given compression rate, a higher MS-SSIM value implies better compression performance. 

\begin{figure*}[t]
	\begin{center}
		\begin{minipage}{0.48\textwidth}
			\centering
			\begin{tabular}{cc}
				{\small FRCI: \emph{\color{red}Street Sign} } & {\small Ours: \emph{\color{blue}Birdhouse}}\\
				\includegraphics[width=0.38\linewidth]{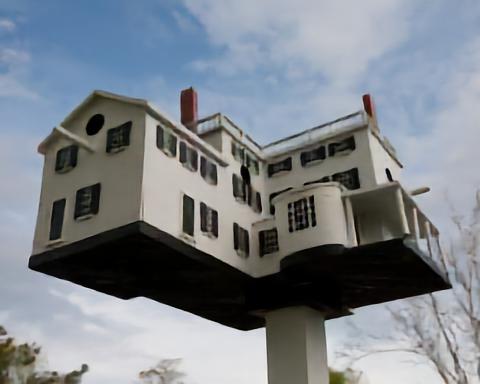}&
				\includegraphics[width=0.38\linewidth]{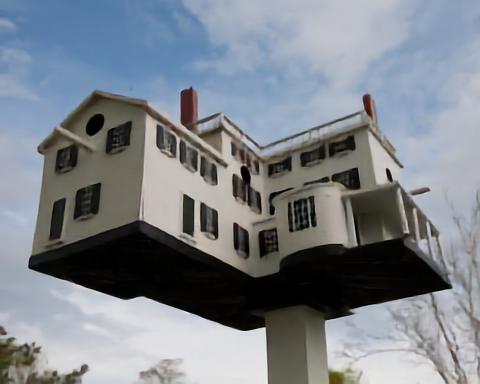}\\
			\end{tabular}
			\caption{Ground-truth: \emph{\color{blue}Birdhouse}.}
			\label{Fig:vs3}
		\end{minipage}
		\begin{minipage}{0.48\textwidth}
			\centering
			\begin{tabular}{cc}
				{\small FRIC: \emph{\color{red}Barbershop}} & {\small Ours: \emph{\color{blue}Trombone}}\\
				\includegraphics[width=0.38\linewidth]{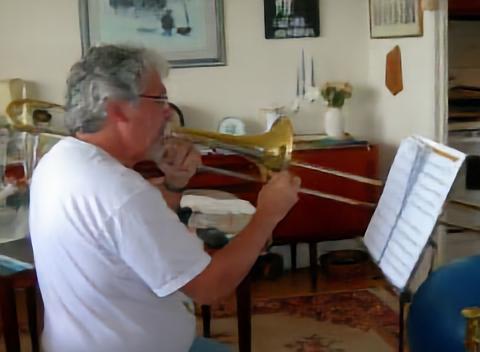}&
				\includegraphics[width=0.38\linewidth]{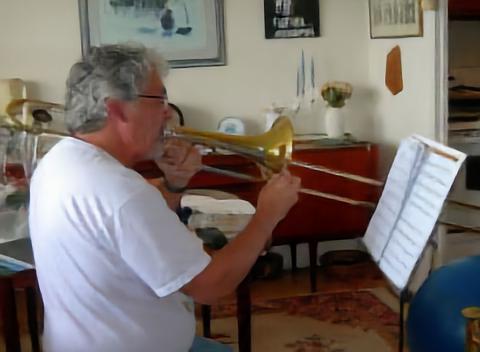}\\
			\end{tabular}
			\caption{Ground-truth: \emph{\color{blue}Trombone}.}
			\label{Fig:vs4}
		\end{minipage}
		
		\begin{minipage}{0.48\textwidth}
			\centering
			\begin{tabular}{cc}
				{\small FRCI: \emph{\color{red}Washbowl} } & {\small Ours: \emph{\color{blue}Pug}}\\
				\includegraphics[width=0.38\linewidth]{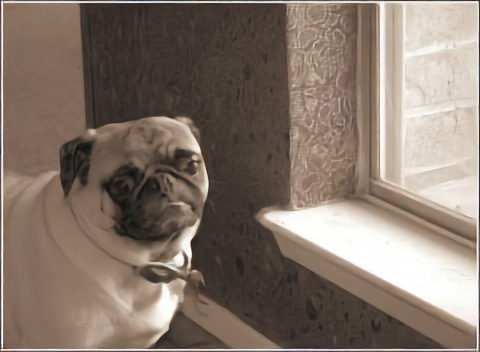}&
				\includegraphics[width=0.38\linewidth]{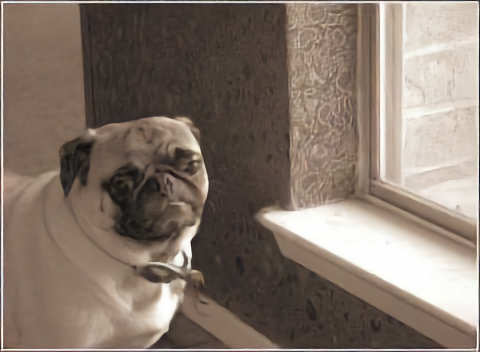}\\
			\end{tabular}
			\caption{Ground-truth: \emph{\color{blue}Birdhouse}.}
		\end{minipage}
		\begin{minipage}{0.48\textwidth}
			\centering
			\begin{tabular}{cc}
				{\small FRIC: \emph{\color{red}Space Bar}} & {\small Ours: \emph{\color{blue}Typewriter}}\\
				\includegraphics[width=0.38\linewidth]{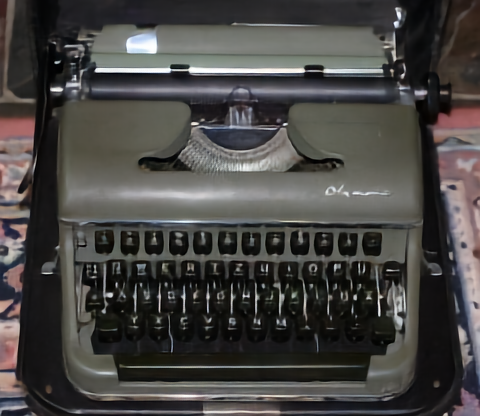}&
				\includegraphics[width=0.38\linewidth]{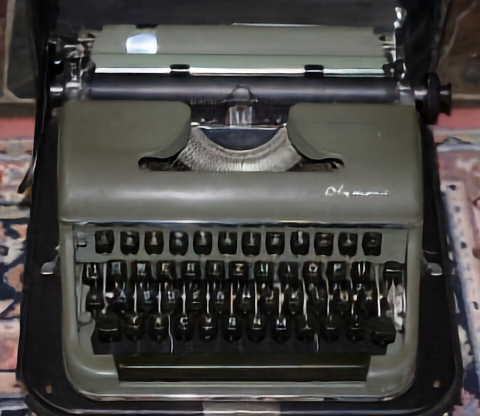}\\
			\end{tabular}
			\caption{Ground-truth: \emph{\color{blue}Trombone}.}
		\end{minipage}
		
		\begin{minipage}{0.48\textwidth}
			\centering
			\begin{tabular}{cc}
				FRCI: \emph{\color{red}Piano} &Ours: \emph{\color{blue}Shoji}\\
				\includegraphics[width=0.38\linewidth]{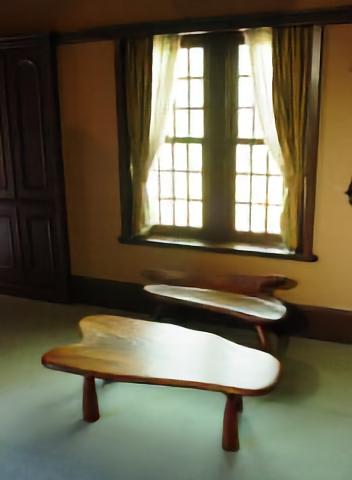}&
				\includegraphics[width=0.38\linewidth]{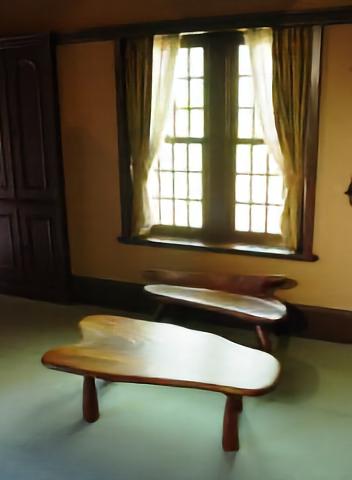}\\
			\end{tabular}
			\caption{Ground-truth: \emph{\color{blue}Shoji}.}
		\end{minipage}
		\begin{minipage}{0.48\textwidth}
			\centering
			\begin{tabular}{cc}
				FRIC: \emph{\color{red} Pointer} & Ours: \emph{\color{blue}Cat}\\
				\includegraphics[width=0.38\linewidth]{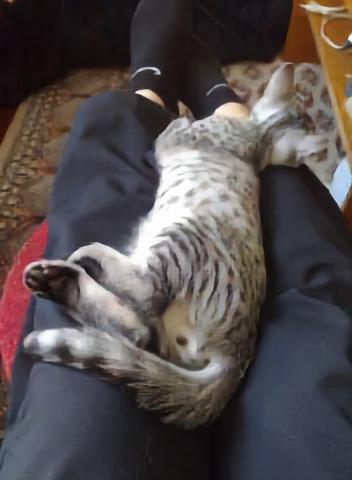}&
				\includegraphics[width=0.38\linewidth]{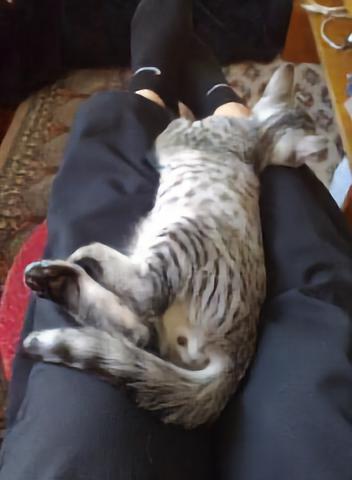}\\
			\end{tabular}
			\caption{Ground-truth: \emph{\color{blue}Cat}.}
			\label{Fig:vs6}
		\end{minipage}
	\end{center}
\end{figure*}

\subsection{Impact of parameters} The proposed image compression method as detailed in Alg.~\ref{Alg:main} has several important parameters: the weight parameter $\lambda$ for reducing the difference between features of original and compressed images similar, $\gamma$ is further used for making features distributions of these images similar which was equal to 1, $u$ was set as 8, and $k$ was set as a series of Gaussian kernels, as analyzed in~\cite{MKMMD}. $b$ is the batch size, which was set as 192 as suggested in~\cite{RNN2}. It is clear that $\lambda$ is the most important parameter for balancing appearance difference evaluated by human eyes and perception difference assessed by machines. Therefore, we first tested the impact of this parameter on the ILSVRC-2012 dataset using the ResNet-18 network as shown Table.~\ref{Tab:comp1}. Wherein, each model is trained on the COCO dataset~\cite{COCO} is as the same as that of the baseline FRIC (Full Resolution Image Compression) method~\cite{RNN2}, and we then employ them on the validation dataset of ILSVRC 2012, respectively. This dataset consists of 50,000 images with different scales and ground-truth labels. All images were first compressed by the proposed DIC, and then recognized by ResNet-18. In addition, the bpp value of original RGB images is 24, and we can achieve a $48\times$ compression rate when bpp = $0.5$, \eg, the file size of a 1\emph{MB} image after compression is about 20\emph{KB}, which is totally enough for recent mobile devices. Therefore, in Table~\ref{Tab:comp1}, bpp values of the conventional method and the proposed method are both equal to $0.5$ for having a better trade-off between image quality and compression rate.

\begin{figure*}[t]
	\begin{center}
		\addtolength{\tabcolsep}{-3pt}
		\begin{tabular}{ccc}
			Original & FRIC & DIC\\
			\includegraphics[width=0.27\linewidth,trim=0 40 0 10,clip]{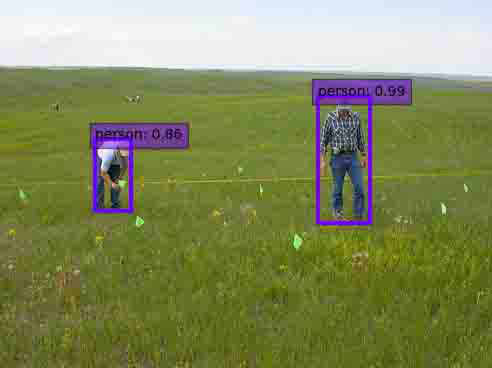}&
			\includegraphics[width=0.27\linewidth,trim=0 40 0 10,clip]{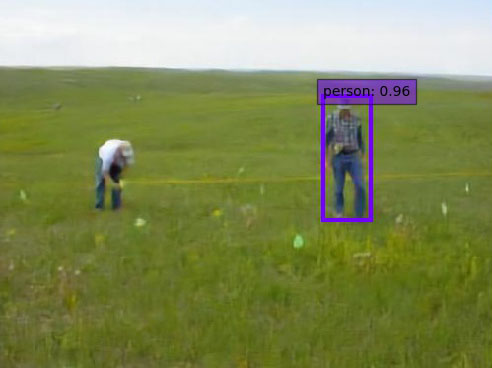}&
			\includegraphics[width=0.27\linewidth,trim=0 40 0 10,clip]{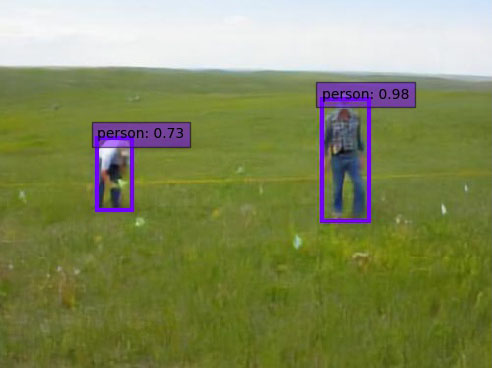}\\
			\includegraphics[width=0.27\linewidth]{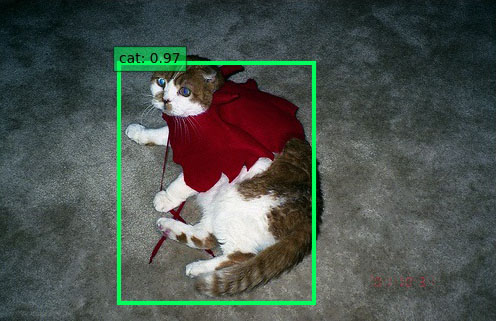}&
			\includegraphics[width=0.27\linewidth]{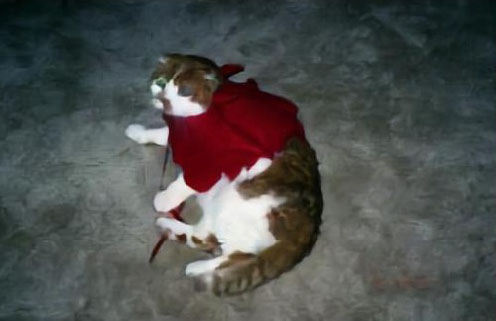}&
			\includegraphics[width=0.27\linewidth]{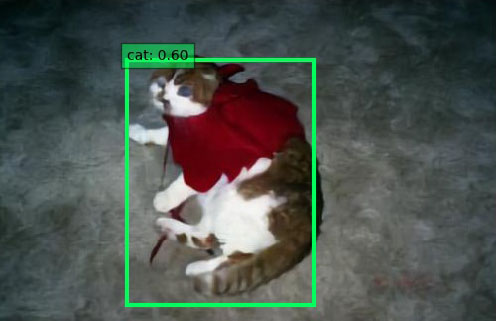}\\
		\end{tabular}
		\caption{Object detection results of images after applying different methods with bpp=0.5.}
		\label{Fig:detection}
	\end{center}
\end{figure*}

\begin{table*}[t]

	\caption{Recognition results on the ILSVRC 2012 dataset, bpp = $0.5$.}
	\label{Tab:comp}
	\begin{center}
		\begin{tabular}{c|c|c|c|c|c|c|c|c|c}
			\hline
			\multirow{2}{2cm}{\centering Method} & \multicolumn{2}{c|}{ResNet-18}& \multicolumn{2}{c|}{ResNet-50}  & {MobileNetV2} & {ShuffleNetV2} & {DenseNet169}  & \multirow{2}{1cm}{\centering MS-SSIM} & \multirow{2}{1cm}{\centering PSNR}\\	
			\cline{2-8}
			&top-1 \emph{acc.} & top-5 \emph{acc.} &top-1 \emph{acc.} & top-5 \emph{acc.} &top-1 \emph{acc.} & top-1 \emph{acc.} & top-1 \emph{acc.} && \\
			\hline\hline
			Original & $69.8\%$ &$89.1\%$& $76.2\%$ &$92.9\%$&          $71.9\%$ &$69.4\%$ &$76.0\%$ & $1.000$ & -\\
			\hline
			JPEG & $62.5\%$ &$84.2\%$& $68.5\%$ &$88.6\%$ &            $59.7\%$ &$58.8\%$&$68.4\%$ & $0.919$  & $24.436$\\
			\hline
			FRIC~\cite{RNN2}& $63.5\%$ & $85.0\%$& $69.2\%$ & $89.0\%$ &           $60.9\%$ &$60.0\%$&$69.3\%$ & $0.921$ & $24.442$\\
			\hline
			DIC w/o MMD & $64.0\%$ & $85.4\%$& $69.6\%$ & $89.1\%$ &             $61.5\%$ &$60.6\%$&$69.7\%$ & $0.925$ & $24.465$\\
			\hline
			DIC w/ MMD & $\textbf{64.3\%}$ & $\textbf{85.5\%}$ & $\textbf{69.8\%}$ & $\textbf{89.2\%}$ &       $\textbf{61.7\%}$ & $\textbf{60.8\%}$ & $\textbf{69.8\%}$  & $0.929$ & $24.471$\\
			\hline
		\end{tabular}
	\end{center}
\end{table*}

It can be found from Table~\ref{Tab:comp1} that a larger $\lambda$ significantly reduces the compression performance, \ie the MS-SSIM values, and a suitable $\lambda$ improves MS-SSIM, since the optimization on visual features can be seen as a powerful regularization for compression approaches considering that $\mathcal{F}(y)=\mathcal{F}(x)$ if the decoder network can generate original images, \ie, $y=x$. When $\lambda = 10^{-6}$, the proposed method can obtain more accurate results than the baseline FRIC. Thus, we kept $\lambda = 10^{-6}$ in the proposed image compression framework, which provides compressed images with better image quality and visual representability. 

\subsection{Comparison Experiments} After investigating the trade-off between the performance of image compression and visual recognition, we then compare the proposed method with state-of-the-art compression methods on the ILSVRC-2012 dataset to verify its effectiveness. All images were first compressed by the proposed DIC (Discernible Images Compression) and several state-of-the-art methods, respectively, and then recognized by ResNet-18. Since the pre-trained network $\mathcal{F}(\cdot)$ in Fcn.~\ref{Fcn:obj} used for extracting features of images before and after compression is part of the ResNet-18, we also employed the ResNet-50,  MobileNetV2, ShuffleNetV2 and DenseNet169 networks to further recognize these images to verify the generalization ability of the proposed method. This ResNet-50 network achieves an $89.2\%$ top-5 accuracy and an $69.8\%$ top-1 accuracy on the ILSVRC 2012 dataset.

Compression results are detailed in Table~\ref{Tab:comp}, where both ResNet-18 and ResNet-50 were pre-trained on the ILSVRC 2012 dataset, and the experiment here aims to investigate how the image compression algorithm affects the subsequent machine learning tasks. Note that, lower bpp values are frequently discussed in many works for obtaining higher compression rates~\cite{lossy3,lossy5}, but compressed images with bpp values lower than $0.5$ have obvious distortions on the compressed image, where results of standard JPEG algorithm are also provided for an explicit comparison. 

It can be found in Table~\ref{Tab:comp}, compressed images generated by all methods downgrade the performance of the subsequent recognition task. It is worth mentioning that, FRIC achieved relatively higher results, since the recurrent network can recover the compressed data iteratively. In contrast, the proposed method can provide compressed images with the highest recognition accuracy on a number of networks. In addition, some recognition results of compressed images are illustrated in Figure~\ref{Fig:vs3}-\ref{Fig:vs6}. Since the proposed image compression method can preserve the perception consistency, images compressed by the proposed method can be recognized, while predictions on FRIC are biased, \eg, a Trombone was recognized as a Barbershop as shown in Figure~\ref{Fig:vs4}. 

 \begin{figure*}[t]
	\begin{center}
		\addtolength{\tabcolsep}{-3pt}
		\begin{tabular}{ccc}
			Original & FRIC & DIC\\
			\includegraphics[width=0.27\linewidth]{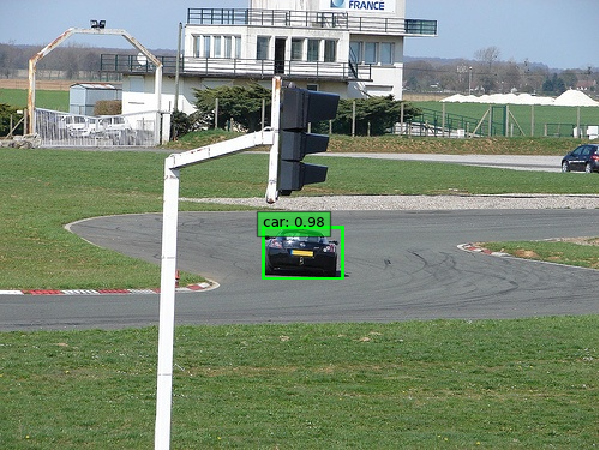}&
			\includegraphics[width=0.27\linewidth]{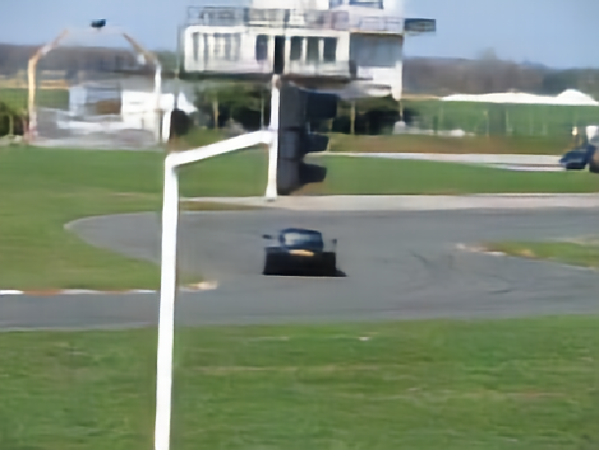}&
			\includegraphics[width=0.27\linewidth]{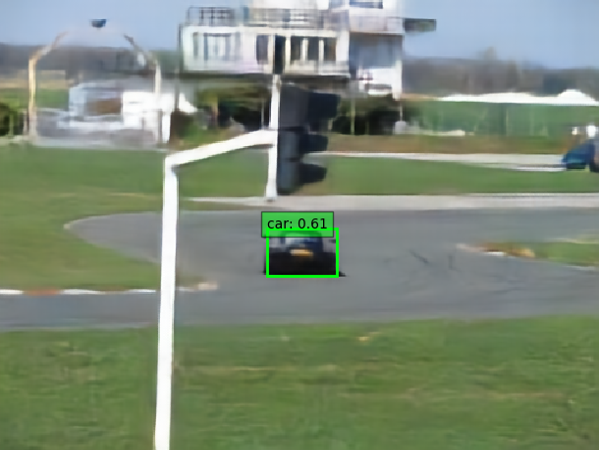}\\
			\includegraphics[width=0.27\linewidth]{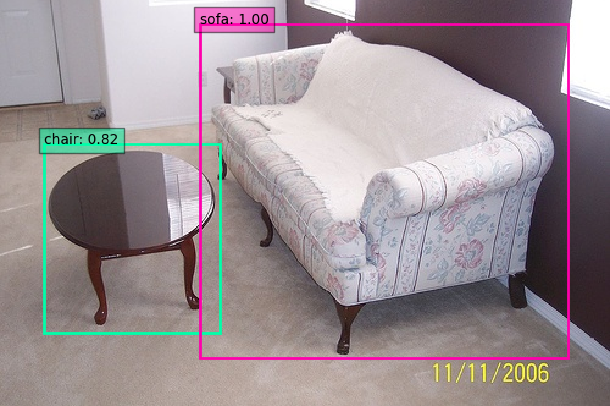}&
			\includegraphics[width=0.27\linewidth]{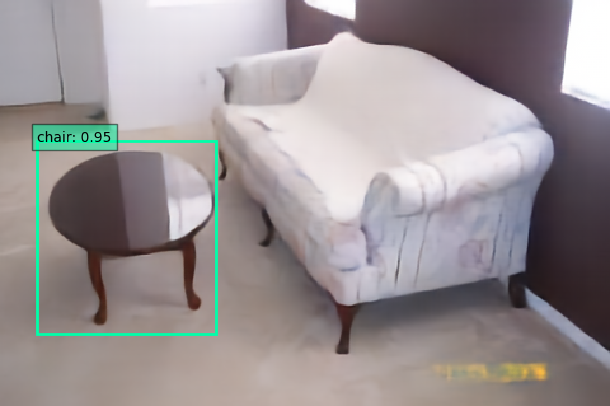}&
			\includegraphics[width=0.27\linewidth]{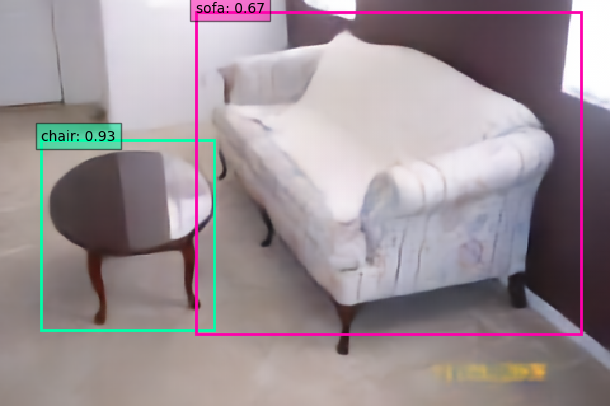}\\

		\end{tabular}
		\caption{Object detection results of images after applying different methods with bpp=0.25.}
		\label{Fig:detection_2}
	\end{center}
\end{figure*}
 
Moreover, we further removed the MMD loss, \ie, the last term in Fcn.~\ref{Fcn:obj}, and re-trained a new model for compressing images to test the impact of the introduced feature distribution regularization. This model was denoted as DIC without MMD (\ie, $\gamma = 0$ in Fcn.~\ref{Fcn:obj}), and recognition results of compressed images using both ResNet-18 and ResNet-50 were reported in Table~\ref{Tab:comp}. The proposed DIC after removing the MMD regularization has an obvious accuracy decline, \eg, its top-1 \emph{acc.} of ResNet-18 is about $0.3\%$ lower than that of the whole DIC method. Since $\mathcal{F}(\cdot)$ converts input images into $512$-dimensional features, and it is very hard to directly minimize differences between these high-dimensional features of original and compressed images. The MMD loss thus can provide a more powerful regularization for obtaining better results.

\begin{table}[t]
	\caption{Object detection results on the VOC 2007 dataset using SSD,  bpp = 0.5, SSD model is trained on VOC07+12.}
	\label{Tab:detection}
	\begin{center}
		\begin{tabular}{c||c|c|c|c}
			\hline
			Method &Original & FRIC & DIC & JPEG \\
			\hline
			\hline
			mAP  & 77.50 & 71.9 & \textbf{72.3} & 71.7 \\
			\hline
			MS-SSIM & 1.000 & 0.937 & 0.940 & 0.935 \\
			\hline
			PSNR & - & 24.671  & 24.694 & 24.655 \\
			\hline
		\end{tabular}
		\vspace{-4pt}
	\end{center}
\end{table}

\subsection{Object Detection after Compressing} Besides the image classification experiment, we further verify the effectiveness of the proposed discernible compressed image generation method on a more complex computer vision application, \ie, object detection. In practice, the deep neural network receives an input image and then outputs locations and labels of each object in the image. Therefore, a slight distortion on the input image could severely damage the prediction results. We selected the SSD (Single Shot MultiBox Detector~\cite{SSD}) trained on the VOC 0712 (Visual Object Classes~\cite{VOC}) as the baseline model to conduct the following experiments.

Table~\ref{Tab:detection} reports the detailed mAP (mean Average Precision) values of original and compressed images (bpp=$0.5$) using different methods, and averaged MS-SSIM results on the VOC 2007 validation set. It is clear that the proposed DIC maintains a higher detection performance. In addition, Figure~\ref{Fig:detection} illustrates some object detection results of original images and compressed images by exploiting the conventional FRIC and the proposed method. It is obvious that the pre-trained SSD model can still detect and recognize objects in compressed images using the proposed DIC algorithm, but cannot accurately recognize those images compressed by exploiting conventional methods. 

In order to further explore the effectiveness of the method, we reduce the bpp of the compressed images to $0.25$, and we visualize the detection results on the compressed images. As shown in Figure~\ref{Fig:detection_2}, in this extreme case, images produced by our proposed DIC can still be correctly detected and retain the semantic information of the image.

\section{Conclusions}\label{Sec:Con}
In this paper, we present a novel deep learning based image compression algorithm. To ensure that the compressed images can also be well recognized by machine learning and deep learning models, we investigate the perceptual consistency. A pre-trained neural network is embedded into the existing encoder-decoder network for compressing natural images, namely Discernible Image Compression (DIC). Beyond directly minimizing the distortion between original images and compressed ones generated by the decoder network, we take the perceptual loss on features into consideration. Compared to state-of-the-art methods, we can generate compressed images of higher image quality by retaining their perception results simultaneously. Experiments on benchmark datasets show that the proposed DIC method can not only produce clearer images of lower storage, but also has limited influence on the downstream visual recognition tasks. The proposed image compression scheme creates a bridge to connect human and machine perceptions. It can be easily deployed for other applications such as denoising, super-resolution, and tracking.

\section*{Acknowledgement}

This work is supported by National Natural Science Foundation of China under Grant No. 61876007, and Australian Research Council under Project DE180101438.

\clearpage
\newpage

{
\bibliographystyle{ACM-Reference-Format}
\bibliography{reference}
}

\end{document}